\documentclass{article}

\usepackage{PRIMEarxiv}
\usepackage{multirow}
\usepackage[utf8]{inputenc} 
\usepackage[T1]{fontenc}    
\usepackage{hyperref}       
\usepackage{url}            
\usepackage{booktabs}       
\usepackage{amsfonts}       
\usepackage{nicefrac}       
\usepackage{microtype}      
\usepackage{lipsum}
\usepackage{fancyhdr}       
\usepackage{graphicx}       
\graphicspath{{media/}}     
\usepackage{natbib}
\usepackage{amsmath}
\title{Buffer replay enhances the robustness of multimodal learning under missing-modality
\thanks{\textit{\underline{Citation}}: 
\textbf{Authors. Title. Pages.... DOI:000000/11111.}} 
}

\author{
  Zhu Hongye, Xuan Liu, Yanwen Ba, Jingye Xue \\
  College of Computer Science and Electronic Engineering \\
  Hunan University \\
  Changsha, China\\
  \texttt{\{Zhuhongye, XuanLiu, YanwenBa, JingyeXue\}@hnu.edu.cn} \\
   \And
  Shigeng Zhang \\
  School of Computer Science  \\
  Central South University \\
  Changsha, China\\
  \texttt{sgzhang@csu.edu.cn} \\
}

\begin{document}
\maketitle

\begin{abstract}
Missing modalities consistently lead to significant performance degradation in multimodal models. Existing approaches either synthesize missing modalities at high computational cost or apply prompt-based fine-tuning that relies only on adjacent-layer features and overlooks long-distance contextual information. 
Such information may offer additional tolerance to errors in scenarios where one or more modalities are missing.
Therefore, REplay Prompting (REP) is introduced. Specifically: (1) modality-wise feature buffers are constructed via a residual bypass to cache early-layer representations and replay them in deeper layers, mitigating information loss as network depth increases; (2) a private–shared feature decoupling strategy is employed, where private buffers preserve fine-grained modality-specific signals while shared buffers encode cross-modal semantics; and (3) a task-aware dynamic initialization mechanism is designed to configure these buffers differently, improving stability and generalization under diverse missing-modality conditions. Experiments on vision–language, vision–language–audio, and temporal multimodal benchmarks show that REP consistently surpasses prior methods under both single- and multi-modality missing conditions, while adding only negligible parameter overhead. These results position REP as a lightweight yet highly effective paradigm for robust multimodal learning in challenging missing-modality scenarios.
Code is available at \href{https://github.com/ZhyHNU/REP}{https://github.com/ZhyHNU/REP}.
\end{abstract}

\keywords{Missing modality \and Prompt fine-tuning \and Multimodal-Learning}

\section{Introduction}
\label{sec:intro}

The rapid development of multimodal learning has given rise to powerful Transformer models, including vision-language models (e.g., \citep{radford2021learning}) and audio-text models (e.g., \citep{guzhov2022audioclip}). These models can effectively leverage information across different modalities and demonstrate excellent performance in tasks such as sentiment analysis \citep{wang2020transmodality, Hu2022UniMSE}, visual recognition \citep{yang2023vid2seq, liu2024visual, liu2025grounding}, and cross-modal retrieval \citep{yang2023vid2seq}. However, missing modalities are common in real-world applications due to factors like sensor malfunctions, privacy constraints, or harsh environments, often making certain modalities unavailable.
\begin{figure}[tbp]
	\centering
	\includegraphics[width=0.5\columnwidth]{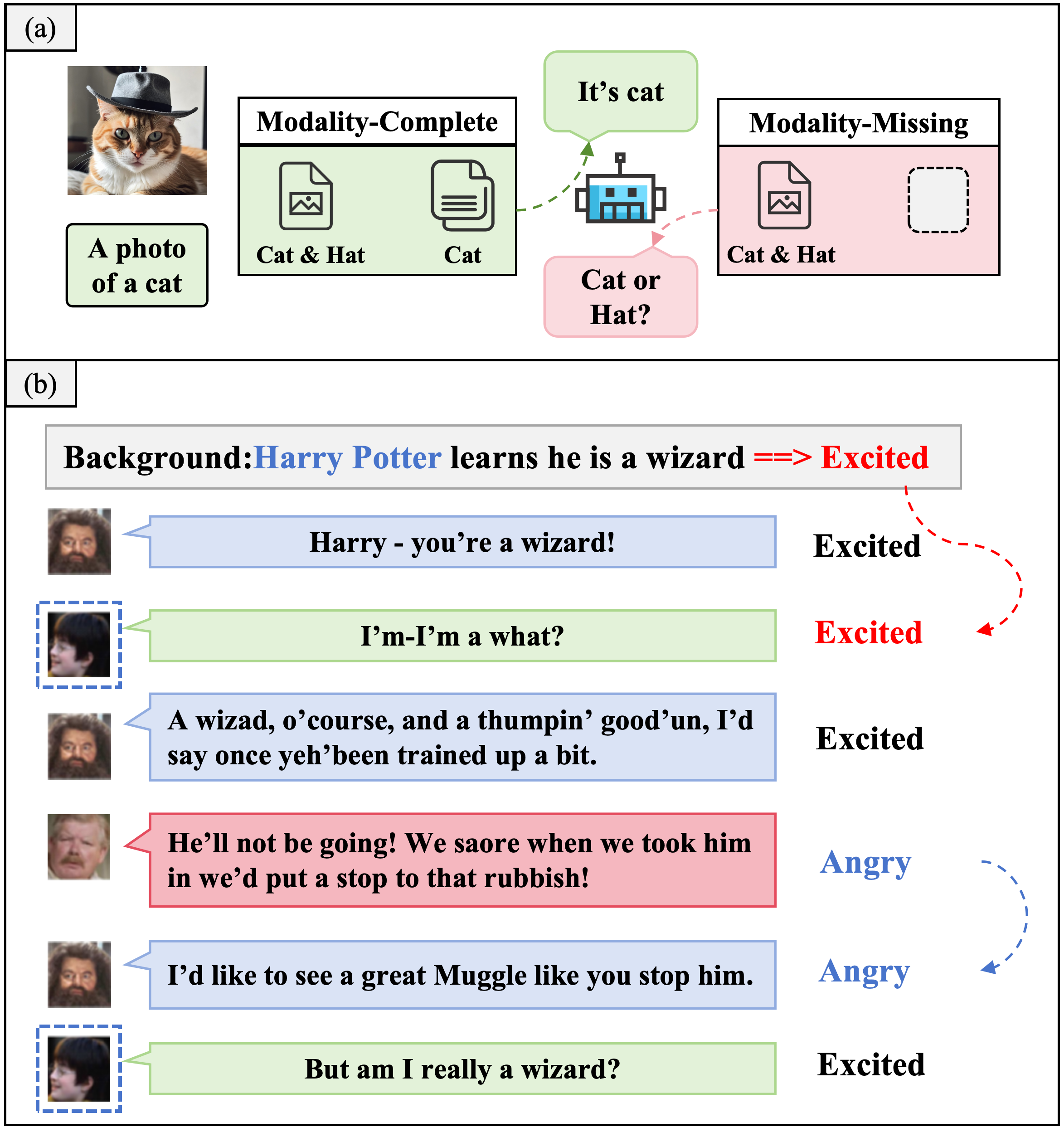}
	\caption{Motivation of the proposed work (a) Modality missing may impair the model’s understanding of the task and its ability to correctly recognize categories.
		(b) Long-distance contextual information is crucial for multimodal emotion recognition; local modeling errors may introduce bias into the global prediction.
		}
	\label{fig1}
\end{figure}

This lack of modality is a significant challenge, as it severely degrades model performance and is a major barrier to the real-world deployment of multimodal technologies. In standard multimodal tasks, redundant information across modalities provides a fault-tolerant buffer. For example, blurry visual features can be compensated by clear textual descriptions. However, in missing-modality scenarios, models often cannot access enough redundant information to fill the gap. As illustrated in Figure \ref{fig1}(a), when the text modality is missing, the model may incorrectly identify the category, significantly affecting its recognition ability and performance \citep{ma2021smil, jang2024towards}.

Modality completion methods rely on generative models to reconstruct missing modalities. Although these approaches have shown progress, they often introduce noise and artificial semantic artifacts during generation, while also incurring considerable computational overhead during both training and inference \citep{wang2023multi, guo2024multimodal}.
In contrast, prompt-based fine-tuning provides a more efficient paradigm. For example, MAP and DCP \citep{lee2023multimodal, hu2024deep} treat missing modalities as prompt inputs and fine-tune the model using mechanisms such as attention prompts or deep-correlation prompts, enabling the model to handle different missing-modality conditions more effectively during inference. However, these methods primarily rely on updates from adjacent-layer features, overlooking the degradation of modality-specific information in early layers. As illustrated in Figure~\ref{fig1}(b), many tasks—such as emotion recognition—depend heavily on long-range contextual information across layers, and partial modality absence can substantially impair global understanding. Furthermore, the compensatory roles of private and shared features under different missing-modality scenarios remain insufficiently modeled.

This paper introduces \textit{Replay Prompting (REP)} for missing modality scenarios. First, we propose a feature caching and replay mechanism. Traditional methods update prompts using only adjacent-layer features. In contrast, our approach preserves early-layer representations through residual bypass feature buffers. By replaying feature buffers into the deeper layers, it effectively alleviates the information decay that occurs with increasing network depth, thereby providing potential information compensation for missing modalities.
Then, we introduce a private-shared feature decoupling strategy. REP adopts a dual-buffer, the private buffer stores modality-specific details, while the shared buffer captures cross-modal information. These buffers are separated using orthogonality constraints and are updated independently. 
Moreover, to further enhance generalization, we introduce a task-aware dynamic initialization that dynamically initializes the buffers based on the task type and input sample, improving generalization and robustness.
REP achieves significant performance improvements across several multimodal benchmarks (covering mainstream modalities such as vision, language, audio, and temporal data) with only a 0.2\% increase in fine-tuning parameters. It consistently achieves state-of-the-art (SOTA) performance in both single-modality and multimodal missing settings, while maintaining an excellent balance between lightweight efficiency and robust high performance.

Overall, the main contributions of this paper include:
\begin{itemize}
	\item Using caching and replay mechanisms to mitigate information decay, providing a fault-tolerant buffer and improving the model's robustness in scenarios with missing modalities.
	
	\item Explicitly modeling private and shared features, capturing modality-specific private information and shared cross-modal information, respectively.
	
	\item The dynamic buffer strategy enhances robustness and generalization.
\end{itemize}

\section{Related work}
\label{sec:relatedwork}

\textbf{Prompt learning} adapts pre-trained models to downstream tasks by introducing learnable prompt tokens into the input space, enabling efficient fine-tuning with minimal parameter updates. Visual Prompt Tuning \cite{jia2022visual} embeds learnable visual prompts into images, while CLIP \cite{radford2021learning} employs hand-crafted text templates to align images and text for strong few-shot performance. CoOp \cite{zhou2022learning} replaces manual templates with learnable text prompts, and CoCoOp \cite{zhou2022conditional} generates prompts conditioned on image features to improve generalization to unseen classes. Progressive Prompts \cite{razdaibiedinaprogressive} further compose task-specific prompts progressively during training, allowing the prompt complexity to adapt over time.

\textbf{Missing modality in multimodal learning} was first introduced and benchmarked by \citet{ma2022multimodal}.  
MAP \cite{lee2023multimodal} treats different missing modalities as independent inputs, but its prompts are initialized with zero matrices, lacking correlation with the inputs.  
DCP \cite{hu2024deep} extends this idea by generating dynamic prompts based on the correlation between input features and prompts; however, its Gaussian-based random generation limits adaptability across diverse scenarios.  
For multimodal sentiment analysis where image, text, or audio modalities may be missing, \citet{guo2024multimodal} introduces generative, missing-signal, and missing-type prompts to handle various missing cases, while \cite{kim2025missing} explores fine-tuning Transformers on small-scale data under the read-only prompt framework \citep{lee2023read} to preserve generalization and stability.  
MaPLe \cite{khattak2023maple} and DePT \citep{zhang2024dept} further enhance cross-task transferability by coupling visual–language prompts and disentangling task-specific from shared knowledge, respectively. LNLN\cite{zhang2024towards} treats language as the dominant modality and enhances sentiment recognition robustness across various random missing and noise scenarios through dominant-modality correction.
In addition, \citep{chen2025i3, li2025multimodal, liu2025hardness} conducted in-depth studies on the problem of incomplete modalities in recommendation systems and multimodal sentiment analysis.

Above methods have been mainly validated on vision-language or limited multimodal tasks with specific modality combinations.
In contrast, we extend our approach to broader multimodal scenarios involving images, text, audio, and sensor signals.
Since existing prompt designs rarely address long-distance modeling under missing modalities, we introduce a replay-based mechanism to better capture long-term and historical dependencies.
We also systematically investigate how different prompt initialization strategies influence model robustness.

\section{Methods}
\subsection{Overview}
\label{3.1}

\begin{figure*}[!htbp]
	\centering
	\includegraphics[width=\linewidth]{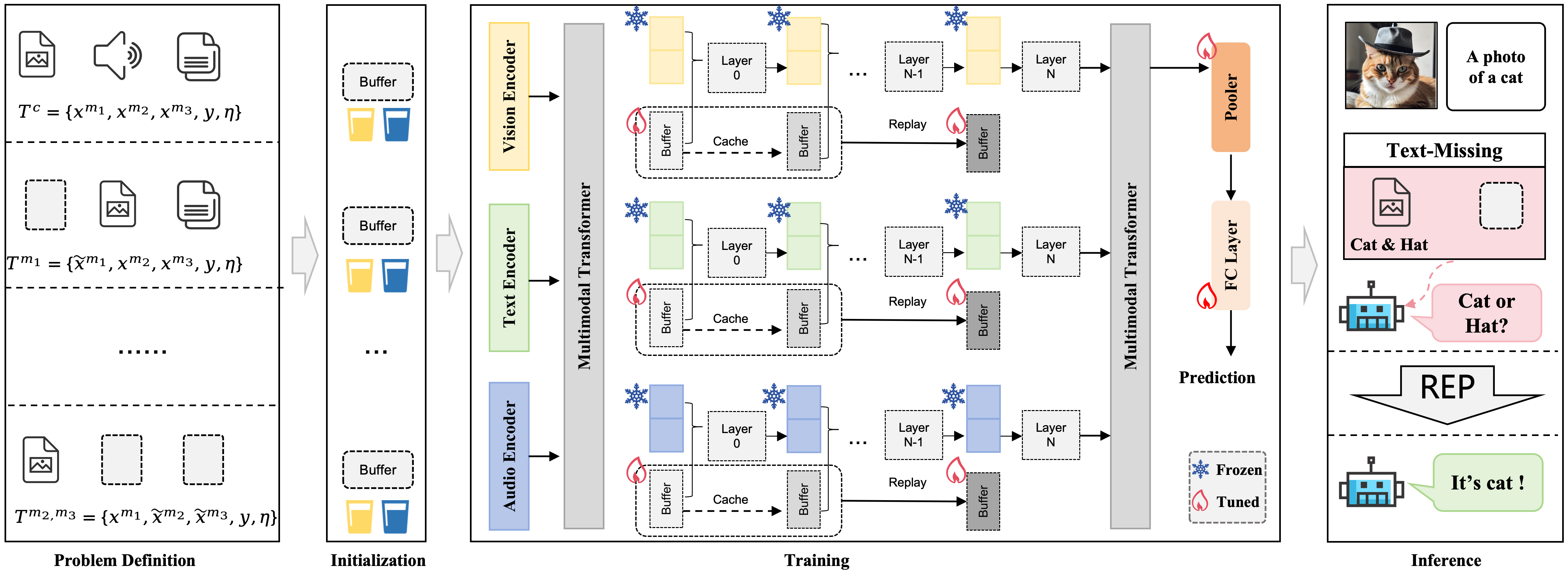} 
	
	\caption{
		Workflow of the proposed REP. 
		(1) Missing scenarios are defined, and different missing types are initialized into private or shared buffers. 
		(2) These buffers cache information from the first $k-1$ layers via residual bypass during fine-tuning, and the cached information is replayed in the $k-th$ layer. 
		(3) During inference, the REP fine-tuned model significantly improves recognition accuracy under missing modality conditions.
	}
	\label{fig:rep_framework}
\end{figure*}

This section introduces REP, a novel multimodal robustness framework designed specifically for modality-missing scenarios. As illustrated in the workflow of Figure \ref{fig:rep_framework}, section \ref{3.2} first presents the definition of modality-missing settings. Following this, section \ref{3.3} introduces the private–shared feature decoupling strategy and explains how the two buffers are cached and updated. Section \ref{3.4} then demonstrates how feature replay is performed at the current layer. Finally, we show the task-aware dynamic initialization strategy in \ref{3.5}, detailing how the two buffers are initialized differently and how this design ultimately enhances generalization.

\subsection{Problem Definition}
\label{3.2}	
We consider a multimodal learning task involving $K$ modalities, whose modality set is defined as:
\begin{equation}
	\mathcal{M}=\{m_1, m_2, \dots, m_K\}.
\end{equation}

For each task, the multimodal representation is formulated as:
\begin{equation}
	T=\{x^{m_1}, x^{m_2}, \dots, x^{m_K}, y, \eta\},
\end{equation}
where $x^{m_i}$ denotes the input feature of modality $m_i$, $y$ is the corresponding label, and $\eta$ is the modality missing rate.

We denote the set of missing modalities as $\mathcal{M}_{\text{miss}} \subseteq \mathcal{M}$, 
and use the modality missing rate $\eta \in [0,1]$ to indicate the proportion of missing information.  
To distinguish different missing cases, we further define:
\begin{itemize}
	\item \textbf{Single-modality missing rate} $\eta_{s}$: the proportion of missing information when only one modality is missing;
	\item \textbf{Multi-modality missing rate} $\eta_{m}$: the total proportion of missing information when multiple modalities are simultaneously missing.
\end{itemize}

In a vision-language task, if the visual modality is missing with a missing rate of $\eta_{s}=70\%$, the sample is represented as:
\begin{equation}
	T^{\mathcal{M}_{\text{miss}}(v)}=\{\tilde{x}^{m_1}, x^{m_2}, y, \eta_{s}=70\%\},
\end{equation}
where $\tilde{x}^{m_1}$ denotes the placeholder input for the missing visual modality.

For example, if both visual and textual modalities are missing, the sample is defined as:
\begin{equation}
	T^{\mathcal{M}_{\text{miss}}(v,l)}=\{\tilde{x}^{m_1}, \tilde{x}^{m_2}, y, \eta_{m}=100\%\},
\end{equation}
where $\eta_{m}=100\%$ indicates that the total missing proportion of both modalities is 100\%, thus the missing rate for each modality is $\frac{\eta_{m}}{|\mathcal{M}_{\text{miss}}|}=50\%$.

\subsection{Private-shared feature decoupling}
\label{3.3}
\begin{figure*}[tbp]
	\centering
	\includegraphics[width=\linewidth]{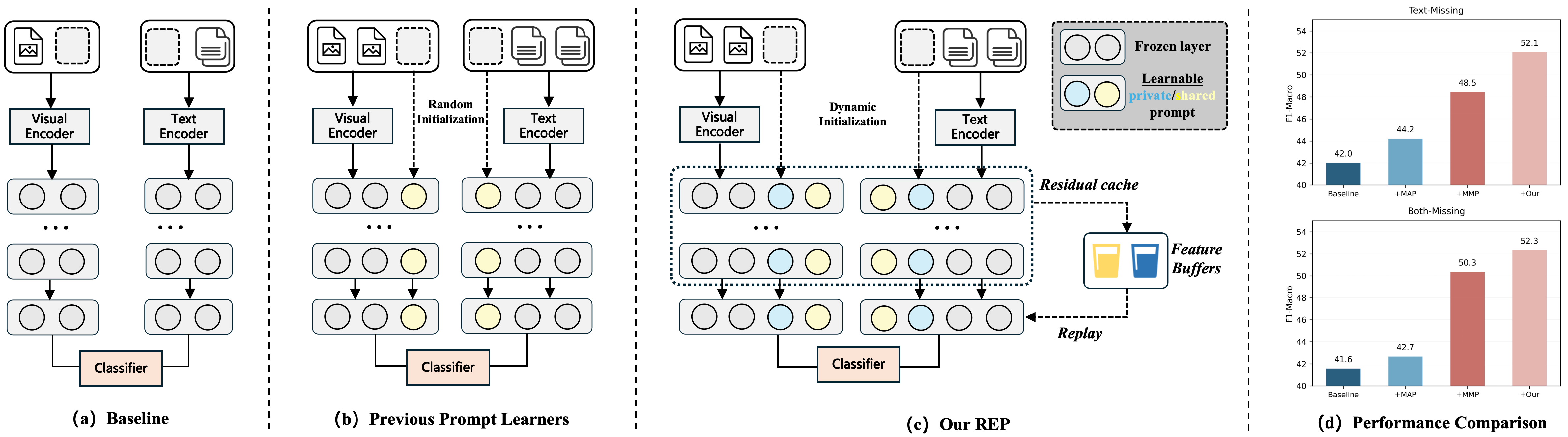} 
	
	\caption{
		(a) Baseline is the CLIP model, using VIT-B/16 as the visual encoder. 
		(b) Previous works such as MAP, MMP, and DCP use random or zero initialization, treating different missing scenarios as independent inputs and fine-tuning the model with prompts. 
		(c) Our proposed REP adopts task-dependent dynamic initialization, caching feature buffers through residual bypass, storing shared and private features separately, and replaying them in deeper layers. 
		(d) Comparison on MM-IMDb, where the missing rate for text-missing is 70\%, and for both-missing, each modality is missing 35\% (total missing rate is also 70\%).
	}
	\label{fig3}
\end{figure*}

Existing multimodal models typically rely on cross-modal shared features for inference tasks. However, when a single modality is missing, modality-specific information may provide potential compensation. Therefore, as shown in Figure \ref{fig3}, we designed a dual-buffer structure to separately capture modality-specific information and cross-modal common semantics.

\subsubsection{Feature Separation}
The private and shared buffers are maintained through independent initialization and update paths, ensuring effective separation of the two feature types. At layer 0, these two kinds of features, together with the original input embeddings, appear in the following form:

\begin{equation}
	\tilde{\mathbf{E}}_m[0] = [\Theta_m(\mathbf{F}_{\text{s}}), \mathbf{F}_{\text{p}}^{m}[0], \mathbf{E}_m[0]]
\end{equation}

Here, $\Theta_m(\mathbf{F}_{\text{s}})$ is the shared feature projected into the modality $m$ space, $\mathbf{F}_{\text{p}}^{m}[0]$ is the modality $m$ private feature, and $\mathbf{E}_m[0]$ is the original input embedding. The three parts are concatenated along the sequence length dimension and input to the Transformer.

To ensure the independence of shared and private features, an orthogonality constraint loss is imposed:

\begin{equation}
	\mathcal{L}_{\text{ortho}} = \sum_{k=1}^{K} \left( \langle \mathbf{F}_{\text{s}}, \mathbf{F}_{\text{p}}^{m}[k] \rangle + \sum_{m \neq m'} \langle \mathbf{F}_{\text{p}}^{m}[k], \mathbf{F}_{\text{p}}^{m'}[k] \rangle \right)
\end{equation}

where $\langle \cdot, \cdot \rangle$ denotes the inner product operation, which measures the similarity between feature vectors. The first term enforces orthogonality between shared and private features, while the second term ensures the orthogonality of private features across different modalities. By minimizing these inner product values, this constraint ensures that the shared space captures cross-modal consistency, while the private space retains modality-specific uniqueness. This constraint is added as a regularization term to the CLIP loss.

\subsubsection{Private Buffer Update}

The private buffer $\mathbf{F}_{\text{p}}^m$ stores modality-specific features that reflect unique characteristics of each modality (e.g., visual textures, semantic patterns, or acoustic cues).  
At each layer $k$, the private buffer is updated through the residual bypass mechanism, which adapts to new features while preserving historical memory:

\begin{equation}
	\mathbf{F}_{\text{p}}^{m}[k] = \alpha_m \cdot G_m(\mathbf{Z}_{\text{p}}^{m}[k]) + (1-\alpha_m) \cdot \mathbf{F}_{\text{p}}^{m}[k-1]
\end{equation}

Here, $\mathbf{Z}_{\text{p}}^{m}[k]$ represents the current features extracted at layer $k$ for modality $m$, processed through normalization and a lightweight transformation network $G_m(\cdot)$ (composed of two layers of MLP and GELU). The coefficient $\alpha_m$ is a learnable scalar that controls the balance between current features and historical memory. A higher $\alpha_m$ places more emphasis on the current layer's features, while a lower value emphasizes accumulated historical information.  
This update mechanism, in the form of residual and bypass storage, preserves low-level details while dynamically integrating new layer semantics, allowing the buffer to continuously accumulate and consolidate modality-specific information.

\subsubsection{Shared Buffer Update}

The shared buffer $\mathbf{F}_{\text{s}}$ stores cross-modal common semantic information. Similar to the private buffer, the shared buffer is also updated through the residual bypass mechanism to maintain semantic consistency and provide compensation in the case of missing modalities:

\begin{equation}
	\mathbf{F}_{\text{s}}[k] = \varepsilon_s \cdot H(\text{C}(\{\mathbf{F}_{\text{p}}^{m}[k]\}_m)) + (1-\varepsilon_s) \cdot \mathbf{F}_{\text{s}}[k-1]
\end{equation}

Here, $\text{C}(\cdot)$ represents the concatenation of the private buffer features from all modalities. $H(\cdot)$ is the shared feature update network (a single-layer MLP with GELU activation) that maps the concatenated features to the shared semantic space. The coefficient $\varepsilon_s$ controls the update rate of the shared features. 

\subsection{Feature Replay}
\label{3.4}
Before entering each Transformer layer, the model performs feature replay by concatenating the private and shared buffer features from earlier layers with the current input features. This process enhances the representation by injecting historical memory into the current layer:

\begin{equation}
	\tilde{\mathbf{E}}_m[k] = \mathbf{E}_m[k] + \beta_{\text{p}} \cdot \mathbf{F}_{\text{p}}^{m}[k-1] + \beta_{\text{s}} \cdot \mathbf{F}_{\text{s}}[k-1]
\end{equation}

Here, $\mathbf{E}_m[k]$ represents the input embedding of modality $m$ at layer $k$, while $\mathbf{F}_{\text{p}}^{m}[k-1]$ and $\mathbf{F}_{\text{s}}[k-1]$ are the private and shared memory features from the previous layer. The weights $\beta_{\text{p}}$ and $\beta_{\text{s}}$ control the amount of historical information injected into the current input.  
This layer-wise replay mechanism allows the model to access both immediate input and historical memory, creating a hierarchical flow of information from shallow to deep layers. The residual bypass replay mechanism strengthens the model's cross-modal reasoning ability and robustness to missing modalities, enabling it to leverage complete historical context.
The aforementioned factors $\alpha_m$, $\varepsilon_s$ and $\beta$ are adjusted adaptively and require no manual tuning.

It is important to note that the depth $d$ of the buffers is determined by the number of layers $k$, while the initial feature embedding length $l$ determines the width of the buffers. These two factors together define the buffer size and the associated parameters and computational costs. In our experiments, we explore the relationship between buffer depth $d$, width $l$, and performance, aiming to maximize performance while minimizing parameter growth.

\subsection{Dynamic Initialization}
\label{3.5}
In the REP method, dynamic initialization initializes feature memory buffers by adding controlled noise to pre-trained modality features, ensuring each modality retains its unique information while introducing sufficient diversity and robustness.

For each modality $m$, the private buffer's initial features are obtained by adding perturbations to the pre-trained embeddings:

\begin{equation}
	\mathbf{F}_{\text{private}}^{m}[0] = \mathcal{E}_m + \epsilon_n \boldsymbol{\zeta}_m, 
	\quad \boldsymbol{\zeta}_m \sim \mathcal{N}(0, \mathbf{I})
\end{equation}

where $\mathbf{F}_{\text{private}}^{m}[0]$ denotes the initial features of the private buffer for modality $m$, with $\mathcal{E}_m$ as the pre-trained embedding, and $\epsilon_n$ controlling the noise intensity. $\boldsymbol{\zeta}_m$ is noise sampled from a standard normal distribution, with independent samples across modalities. This process introduces controlled diversity to each modality's initial representation, improving generalization.

The shared buffer is initialized with normalized Gaussian noise to provide a flexible semantic space for cross-modal collaboration:

\begin{equation}
	\mathbf{F}_{\text{shared}}[0] = \frac{\boldsymbol{\xi}}{\|\boldsymbol{\xi}\|_2}, 
	\quad \boldsymbol{\xi} \sim \mathcal{N}(0, \mathbf{I})
\end{equation}

where $\mathbf{F}_{\text{shared}}[0]$ stores cross-modal common semantic information. $\boldsymbol{\xi}$ is a random vector sampled from the standard normal distribution, with the normalization ensuring the vector lies on the unit hypersphere. This initialization strategy enables flexible cross-modal collaboration while avoiding numerical instability.

This dynamic initialization strategy ensures diversified initial features for each modality while maintaining separation between shared and private information. In the experiments, we analyze the impact of $\epsilon_n$ and different noise types on performance.

\section{Experiments}
\begin{table*}[htbp]
	\centering
	\caption{Comparison of performance on MM-IMDb, Food101, and Hateful Memes datasets under different missing modality rates $\eta$. 
		Bold values indicate the best performance.}
	\label{tab:main_results}
	\vspace{0.5em}
	\resizebox{\textwidth}{!}{
		\setlength{\tabcolsep}{3pt}
		\renewcommand{\arraystretch}{0.9}
		\begin{tabular}{c|c|cc|cccccccc}
			\toprule
			\multirow{2}{*}{Dataset} & 
			\multirow{2}{*}{\shortstack{Missing \\ rate $\eta$}} &
			\multicolumn{2}{c|}{Train / Test} &
			\multicolumn{8}{c}{Methods} \\
			\cmidrule(lr){3-4} \cmidrule(lr){5-12}
			& & Image & Text 
			& Baseline 
			& CoOp~\cite{zhou2022learning}
			& MAP~\cite{lee2023multimodal}
			& MaPLe~\cite{khattak2023maple}
			& DePT~\cite{zhang2024dept} 
			& MMP~\cite{kim2024missing} 
			& DCP~\cite{hu2024deep} 
			& \textbf{Ours} \\
			\midrule
			
			\multirow{9}{*}{\shortstack{MM-IMDb\\(F1-Macro)}} 
			
			&  & 100\% & 50\% 
			& 44.73 & 48.06 & 48.88 & 49.58 & 50.64 & 49.21 & 52.13 & \textbf{54.05} \\
			
			& 50\% & 50\% & 100\% 
			& 45.96 & 49.89 & 51.46 & 52.32 & 52.78 & 53.67 & 54.32 & \textbf{55.40} \\
			
			&  & 75\% & 75\% 
			& 44.12 & 48.37 & 49.32 & 49.56 & 50.87 & 51.09 & 52.46 & \textbf{55.27} \\
			\cmidrule(lr){2-12}
			
			&  & 100\% & 30\% 
			& 42.02 & 44.13 & 44.22 & 45.52 & 46.38 & 48.46 & 48.52 & \textbf{52.08} \\
			
			& 70\% & 30\% & 100\% 
			& 42.95 & 46.88 & 46.30 & 50.64 & 52.13 & 52.62 & 53.14 & \textbf{53.71} \\
			
			&  & 65\% & 65\% 
			& 41.64 & 46.84 & 42.66 & 48.30 & 50.32 & 50.34 & 51.69 & \textbf{52.31} \\            
			\cmidrule(lr){2-12}
			
			&  & 100\% & 10\% 
			& 40.31 & 44.06 & 45.32 & 46.84 & 47.56 & 42.07 & 49.26 & \textbf{50.05} \\
			
			& 90\% & 10\% & 100\% 
			& 46.02 & 49.89 & 50.13 & 50.88 & 51.22 & 51.23 & 51.95 & \textbf{52.05} \\
			
			&  & 55\% & 55\% 
			& 40.18 & 44.12 & 44.87 & 45.12 & 46.54 & 47.81 & 48.44 & \textbf{50.76} \\
			
			\midrule
			
			\multirow{9}{*}{\shortstack{Food101\\(Accuracy)}} 
			
			&  & 100\% & 50\% 
			& 72.18 & 77.45 & 77.89 & 79.64 & 80.16 & 78.81 & 82.11 & \textbf{83.65} \\
			
			& 50\% & 50\% & 100\% 
			& 80.16 & 87.02 & 87.16 & 87.35 & 82.14 & 86.90 & 88.29 & \textbf{88.56} \\
			
			&  & 75\% & 75\% 
			& 75.49 & 81.82 & 81.72 & 82.34 & 83.12 & 82.35 & 85.24 & \textbf{85.55} \\
			\cmidrule(lr){2-12}
			
			&  & 100\% & 30\% 
			& 70.42 & 76.34 & 74.53 & 77.02 & 77.34 & 75.41 & 78.87 & \textbf{80.40} \\
			
			& 70\% & 30\% & 100\% 
			& 78.11 & 84.78 & 86.18 & 85.89 & 86.12 & 87.11 & \textbf{87.32} & 87.22 \\
			
			&  & 65\% & 65\% 
			& 73.02 & 78.87 & 79.08 & 79.84 & 81.46 & 81.22 & 81.87 & \textbf{82.68} \\
			\cmidrule(lr){2-12}
			
			&  & 100\% & 10\% 
			& 66.41 & 71.87 & 73.14 & 73.46 & 74.12 & 78.13 & 75.26 & \textbf{77.86} \\
			
			& 90\% & 10\% & 100\% 
			& 76.68 & 83.12 & 83.96 & 83.12 & 83.56 & 84.35 & 85.78 & \textbf{85.84} \\
			
			&  & 55\% & 55\% 
			& 70.31 & 76.46 & 76.58 & 77.85 & 78.12 & 79.67 & 79.87 & \textbf{80.33} \\
			
			\midrule
			
			\multirow{9}{*}{\shortstack{Hateful Memes\\(AUROC)}} 
			
			&  & 100\% & 50\% 
			& 55.77 & 60.56 & 60.31 & 60.87 & 61.87 & 62.12 & 62.32 & \textbf{69.92} \\
			
			& 50\% & 50\% & 100\% 
			& 57.62 & 62.41 & 62.35 & 63.13 & 63.88 & 64.01 & 64.46 & \textbf{66.23} \\
			
			&  & 75\% & 75\% 
			& 59.83 & 64.87 & 65.84 & 65.46 & 65.86 & 65.42 & 66.02 & \textbf{67.44} \\
			\cmidrule(lr){2-12}
			
			&  & 100\% & 30\% 
			& 57.44 & 62.60 & 59.11 & 63.14 & 63.84 & 61.39 & 64.12 & \textbf{70.01} \\
			
			& 70\% & 30\% & 100\% 
			& 58.71 & 64.07 & 63.06 & 64.12 & 64.54 & 65.24 & 65.48 & \textbf{66.17} \\
			
			&  & 65\% & 65\% 
			& 59.54 & 64.82 & 66.07 & 65.23 & 65.48 & 65.17 & 66.08 & \textbf{66.89} \\
			\cmidrule(lr){2-12}
			
			&  & 100\% & 10\% 
			& 55.29 & 60.03 & 57.21 & 60.71 & 61.14 & 62.06 & 62.08 & \textbf{68.92} \\
			
			& 90\% & 10\% & 100\% 
			& 56.92 & 61.46 & 61.52 & 61.87 & 62.42 & 61.97 & 63.57 & \textbf{63.87} \\
			
			&  & 55\% & 55\% 
			& 57.41 & 62.93 & 63.34 & 63.68 & 63.12 & 64.27 & 64.58 & \textbf{64.77} \\
			
			\bottomrule
		\end{tabular}
	}
\end{table*}

\subsection{Implementation Details}
\subsubsection{Benchmark}

We construct a systematic benchmark covering two multimodal task settings: 
vision-language (VL), vision-language-audio (VLA) and acoustic–seismic (AS).

\textbf{VL Benchmark}  includes three datasets:
MM-IMDb (F1-Macro)~\cite{arevalo2017gated}, 
UPMC Food-101 (accuracy)\cite{bossard14}, 
and Hateful Memes (AUROC)~\cite{kiela2020hateful}. 
These datasets span multimodal sentiment analysis, food image--text recognition, 
and cross-modal sarcasm detection. 
The baseline model is CLIP~\cite{radford2021learning}, 
where the image encoder is ViT-B/16, 
and we directly use publicly released pre-trained weights.

\textbf{VLA Benchmark}  is conducted on CH-SIMS~\cite{yu2020ch} 
and MOSI~\cite{zadeh2016mosi}, using ACC and F1 as evaluation metrics. 
These datasets cover vision--text--audio multimodal sentiment analysis. 
The baseline model is MuLT~\cite{tsai2019MULT}, pre-trained on the CMU-MOSEI dataset~\cite{zadeh2018multimodal} without prompts.

\textbf{AS Benchmark} is evaluated on the MOD dataset\cite{liu2023focal}, which is used for mobile objects (vehicle, human) classification. And model is pre-trained on the MOD pre-training dataset.

\subsubsection{Missing Settings}
We use fixed placeholder inputs to simulate unavailable modalities while maintaining consistent input structure:

\begin{itemize}
	\item \textbf{Text-missing}: Empty string $\rightarrow$ minimal valid text embedding via tokenizer
	\item \textbf{Image-missing}: All-one pixel image $\rightarrow$ semantics-free visual representation via patch embedding
	\item \textbf{Audio-missing}: Fixed-length all-zero vector $\rightarrow$ minimal semantic audio embedding via front-end encoder
\end{itemize}

\begin{table*}[t]
	\centering
	\caption{Comparison on \textbf{CH-SIMS}, \textbf{MOSEI}, and \textbf{MOSI} datasets under missing rate $\eta = 70\%$. 
		[Audio-only] indicates that the other two modalities are missing, each with a missing rate of 35\%.
		[Audio-Missing] indicates that only the audio modality is missing, with a missing rate of 70\%. Bold values indicate the best performance.}
	\label{tab:multimodal_all}
	\vspace{0.5em}
	\resizebox{\textwidth}{!}{
		\scriptsize
		\setlength{\tabcolsep}{2pt}
		\renewcommand{\arraystretch}{0.9}
		\begin{tabular}{c|c|cccccccccccc|cc}
			\toprule
			\multirow{2}{*}{Dataset} & \multirow{2}{*}{Method} 
			& \multicolumn{2}{c}{Audio-only} 
			& \multicolumn{2}{c}{Image-only} 
			& \multicolumn{2}{c}{Text-only} 
			& \multicolumn{2}{c}{Text-missing} 
			& \multicolumn{2}{c}{Image-missing} 
			& \multicolumn{2}{c}{Audio-missing} 
			& \multicolumn{2}{c}{Avg.} \\
			\cmidrule(lr){3-16}
			& & Acc & F1 & Acc & F1 & Acc & F1 & Acc & F1 & Acc & F1 & Acc & F1 & Acc & F1 \\
			\midrule
			\multirow{4}{*}{\shortstack{CH-SIMS}} 
			& MMIN & 65.21 & 77.09 & 64.28 & 73.36 & 78.91 & 78.67 & 77.32 & 77.33 & 77.40 & 77.48 & 77.40 & 77.48 & 73.42 & 76.90 \\
			& LNLN & 64.98 & 76.41 & 64.01 & 73.47 & 78.56 & 78.65 & 77.11 & 77.20 & 77.51 & 77.47 & 77.51 & 77.47 & 73.28 & 76.78 \\
			& MPLMM & 65.93 & 77.10 & 65.28 & 74.02 & 79.75 & 78.74 & 77.45 & 77.84 & 77.97 & 77.95 & 77.97 & 77.95 & 74.06 & 77.27 \\
			& \textbf{REP} & \textbf{67.19} & \textbf{78.22} & \textbf{66.62} & \textbf{75.81} & \textbf{81.09} & \textbf{80.08} & \textbf{78.57} & \textbf{79.29} & \textbf{79.31} & \textbf{79.29} & \textbf{79.31} & \textbf{79.29} & \textbf{75.35} & \textbf{78.66} \\
			\midrule
			\multirow{4}{*}{\shortstack{MOSEI}} 
			& MMIN & 67.11 & 68.67 & 68.17 & 69.74 & 78.67 & 78.71 & 79.94 & 79.96 & 79.32 & 79.29 & 79.32 & 79.29 & 75.42 & 75.94 \\
			& LNLN & 66.94 & 68.74 & 68.11 & 69.79 & 78.21 & 78.30 & 79.41 & 79.47 & 79.63 & 79.71 & 79.63 & 79.71 & 75.32 & 75.95 \\
			& MPLMM & 67.33 & 68.71 & 68.21 & 69.91 & 79.12 & 79.17 & 80.45 & 80.43 & 80.11 & 80.13 & 80.11 & 80.13 & 75.89 & 76.41 \\
			& \textbf{REP} & \textbf{68.58} & \textbf{70.05} & \textbf{69.98} & \textbf{71.25} & \textbf{80.46} & \textbf{80.51} & \textbf{81.68} & \textbf{81.77} & \textbf{81.33} & \textbf{81.47} & \textbf{81.33} & \textbf{81.47} & \textbf{77.23} & \textbf{77.75} \\
			\midrule
			\multirow{4}{*}{\shortstack{MOSI}} 
			& MMIN & 59.16 & 60.12 & 61.01 & 61.98 & 80.10 & 80.16 & 63.79 & 64.08 & 80.50 & 80.33 & 80.46 & 80.63 & 70.84 & 71.22 \\
			& LNLN & 62.26 & 62.35 & 61.63 & 62.12 & 79.81 & 80.10 & 64.54 & 64.33 & 79.89 & 79.84 & 80.74 & 80.93 & 70.48 & 70.11 \\
			& MPLMM & 62.71 & 63.65 & 63.12 & 63.74 & 80.12 & 80.31 & 65.02 & 65.41 & 80.76 & 81.09 & 81.12 & 81.19 & 72.14 & 72.57 \\
			& \textbf{REP} & \textbf{64.25} & \textbf{65.28} & \textbf{64.86} & \textbf{65.41} & \textbf{81.43} & \textbf{81.52} & \textbf{66.38} & \textbf{66.79} & \textbf{81.72} & \textbf{81.83} & \textbf{81.72} & \textbf{81.83} & \textbf{73.73} & \textbf{73.78} \\
			
			\bottomrule
		\end{tabular}
	}
\end{table*}

\subsection{Results on VL tasks}

To validate the robustness and generalization capabilities of our proposed method, we introduce different modality missing rates and various missing scenarios across three VL datasets. For example, missing rate $\eta=70\%$ corresponds to severe missing conditions, including image-only missing 70\%, text-only missing 70\%, or both modalities missing 35\% each. The comparison results are shown in Table 1, with the best results highlighted in bold.

REP demonstrates consistent performance advantages across all 27 experimental configurations. 
Its superiority is particularly evident on the Hateful Memes dataset, where it achieves the best results in all nine missing-modality settings, outperforming the second-best method DCP by an average of 4.5 percentage points. 
Under severe missing conditions ($\eta = 70\%$), REP attains 52.08 F1-Macro on MM-IMDb and 70.01 AUROC on Hateful Memes, corresponding to improvements of 7.3\% and 9.2\%, respectively. 
In extreme missing scenarios ($\eta = 90\%$), where either the text or image modality is nearly absent, the advantage of REP becomes even more pronounced. 
For example, under the text-severely-missing condition on Hateful Memes, REP surpasses DCP by 11\%, indicating that the cross-layer feature buffering and replay mechanism effectively leverages historical information from complete modalities to compensate for missing signals, thereby providing greater fault tolerance and significantly improving prediction stability. 
Overall, REP exhibits consistent benefits across various missing-modality patterns and datasets, validating the effectiveness of its feature buffering and replay strategy in mitigating information degradation and enhancing robustness under modality-incomplete conditions.

\subsection{Results on VLA tasks}

We adopt a pre-training and prompt-tuning pipeline using a frozen MulT backbone \cite{tsai2019MULT}, trained with random modality missing to enhance robustness. Experiments are conducted on CH-SIMS, MOSEI, and MOSI, and compared with MMIN \cite{zhao2021missing}, LNLN \cite{zhang2024towards}, and MPLMM \cite{guo2024multimodal}. As shown in Table~\ref{tab:multimodal_all}, REP consistently achieves the best performance across all single-modality missing scenarios.

REP performs strongly in dual-modality missing settings such as Audio-only and Image-only. When text is the only remaining modality, REP achieves the highest accuracy on all benchmarks, reaching 81.09\% on CH-SIMS and 80.46\% on MOSEI. Across all missing-modality conditions, REP delivers the best overall results on the three datasets, with F1 scores of 78.66 on CH-SIMS and 77.75 on MOSEI. These findings show that the cross-layer replay mechanism provides effective compensation for missing and enables performance in multimodal sentiment recognition.

\subsection{Resuls on AS task}
The MOD dataset consists of acoustic (8kHz) and seismic (100Hz) signals collected for vehicle and pedestrian classification, covering a total of seven classes. We evaluate the proposed REP method on this dataset, focusing on its performance on time-series sensor modalities. The comparison methods include Swin Transformer (Baseline) and FOCAL, and the results are shown in the figure below:
\begin{figure}[htbp]
	\centering
	\includegraphics[width=\columnwidth]{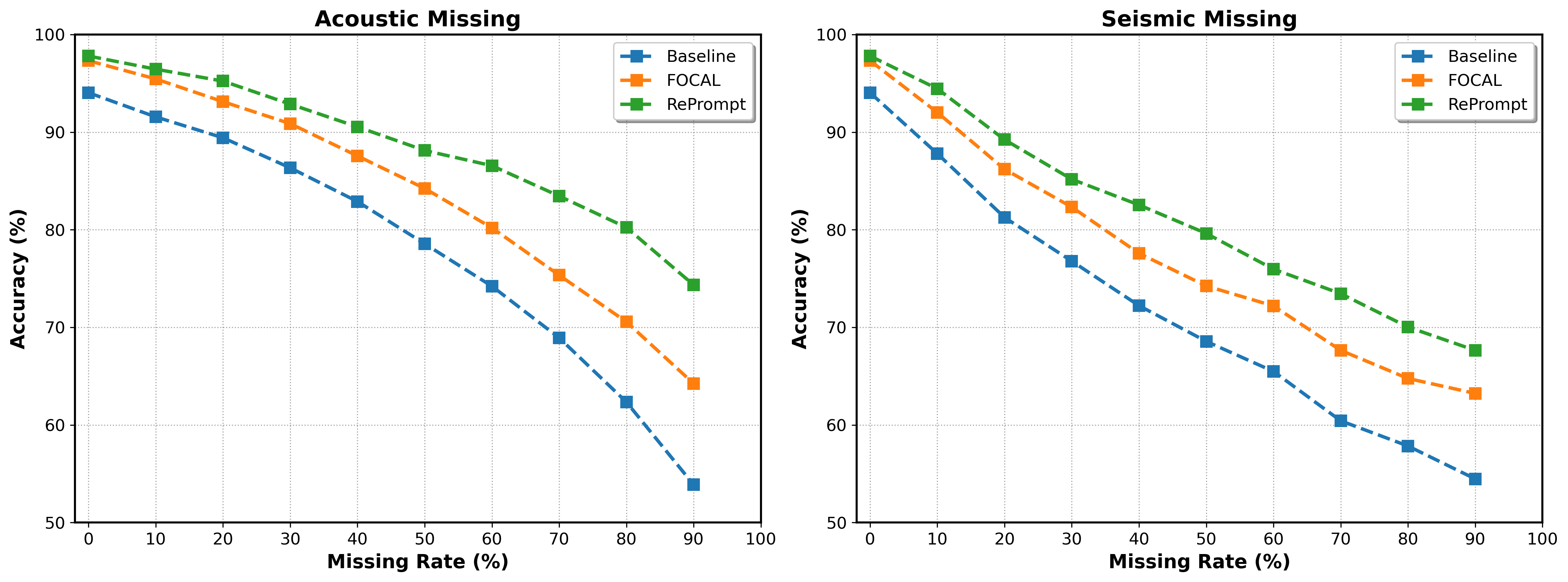}
	\caption{Modal missing tests on acoustic-seismic temporal sensor modalities.}
	\label{fig4}
\end{figure}

Experimental results show that different vehicles produce significantly different ground vibration patterns, and the absence of this modality leads to a substantial drop in accuracy, especially when the missing rate exceeds 50\%. Methods without prompt-based adaptation suffer from a larger performance decline in such scenarios. Our proposed REP effectively mitigates the performance degradation caused by missing time-series sensor signals through feature replay and information compensation, demonstrating stronger robustness across all missing rates as well as the potential to be extended to more modalities.

\subsection{Ablation study}

In this section, we will conduct ablation studies to investigate the following aspects: (1) the impact of noise ratios/types in initialization, (2) the impact of the replay mechanism, (3) the impact of private–shared feature separation,and
(4) how buffer configurations influence computational cost, parameters, and overall performance.

\textbf{Module ablation.} 
To evaluate the effectiveness of each component, we conduct ablation experiments on the Food101 dataset under a missing-modality rate of $\eta = 70\%$. As shown in Table~\ref{tab:food101_ablation}, we progressively add dynamic initialization, private–shared feature separation, and the replay mechanism on top of the CLIP baseline to examine their individual contributions. Furthermore, in Table~\ref{tab:food101_ablation_part2}, we investigate the additional effects of removing the private or shared buffer, as well as the impact of different noise types used during initialization.

\begin{table}[htbp]
	\centering
	\caption{Ablation study on main components of REP($\eta = 70\%$). }
	\label{tab:food101_ablation}
	\renewcommand{\arraystretch}{1.05}
	\setlength{\tabcolsep}{5pt}
	\resizebox{0.65\linewidth}{!}{
		\begin{tabular}{l|ccc}
			\toprule
			\textbf{Method Variant} 
			& \textbf{Text-missing} 
			& \textbf{Image-missing} 
			& \textbf{Both-missing} \\ 
			\midrule
			Baseline 
			& 70.42 & 78.00 & 73.02 \\
			\midrule
			+ Dynamic Initialization 
			& 73.11 & 80.42 & 75.48 \\
			
			+ Dual buffers
			& 76.84 & 83.91 & 78.91 \\
			
			+ Replay (Full REP)
			& \textbf{80.40} & \textbf{87.32} & \textbf{82.68} \\

			\bottomrule
		\end{tabular}
	}
\end{table}

\begin{table}[ht]
	\centering
	\caption{Ablation of buffers and noise types ($\eta = 70\%$).}
	\label{tab:food101_ablation_part2}
	\renewcommand{\arraystretch}{1.05}
	\setlength{\tabcolsep}{6pt}
	\resizebox{0.65\linewidth}{!}{
		\begin{tabular}{l|ccc}
			\toprule
			\textbf{Method Variant} 
			& \textbf{Text-missing} 
			& \textbf{Image-missing} 
			& \textbf{Both-missing} \\ 
			\midrule
			\textbf{Full REP} 
			& \textbf{80.40} & \textbf{87.22} & \textbf{82.68} \\
			\midrule
	
			w/o Private Buffer 
			& 78.74 & 85.92 & 81.05 \\
			Gaussian $\rightarrow$ Uniform 
			& 79.11 & 85.36 & 81.08 \\
			Gaussian $\rightarrow$ Laplace 
			& 78.85 & 85.92 & 81.14 \\
			\bottomrule
		\end{tabular}
	}
\end{table}
\begin{figure*}[t]
	\centering
	\includegraphics[width=0.9\textwidth]{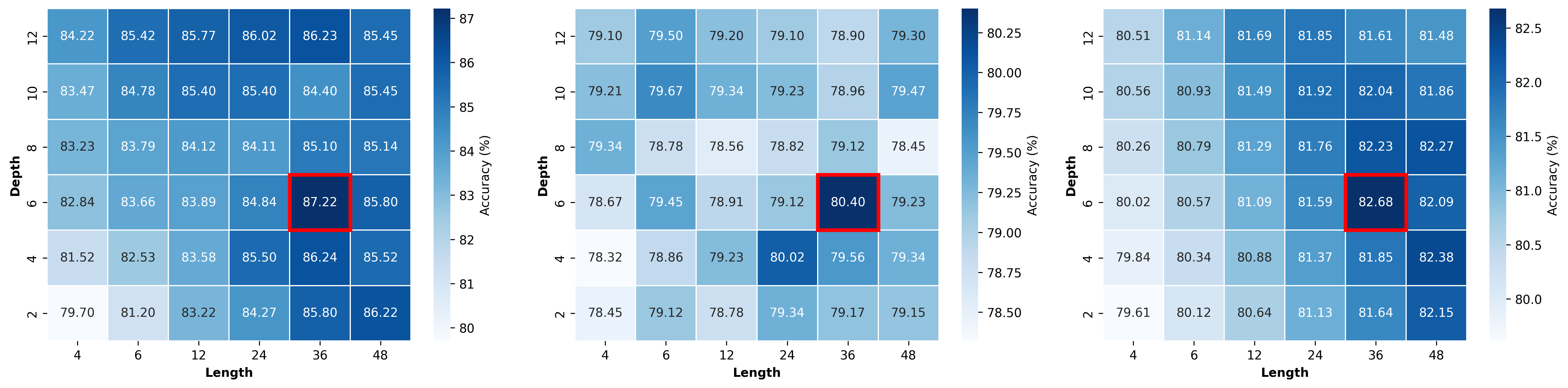}
	\caption{Ablation study on REP buffer configuration. Results are under 70\% missing rate. From left to right are the text-missing, image-missing, and both-missing settings, respectively.
	} 
	\label{fig5}
\end{figure*}

\begin{figure}[htbp]
	\centering
	\includegraphics[width=0.7\columnwidth]{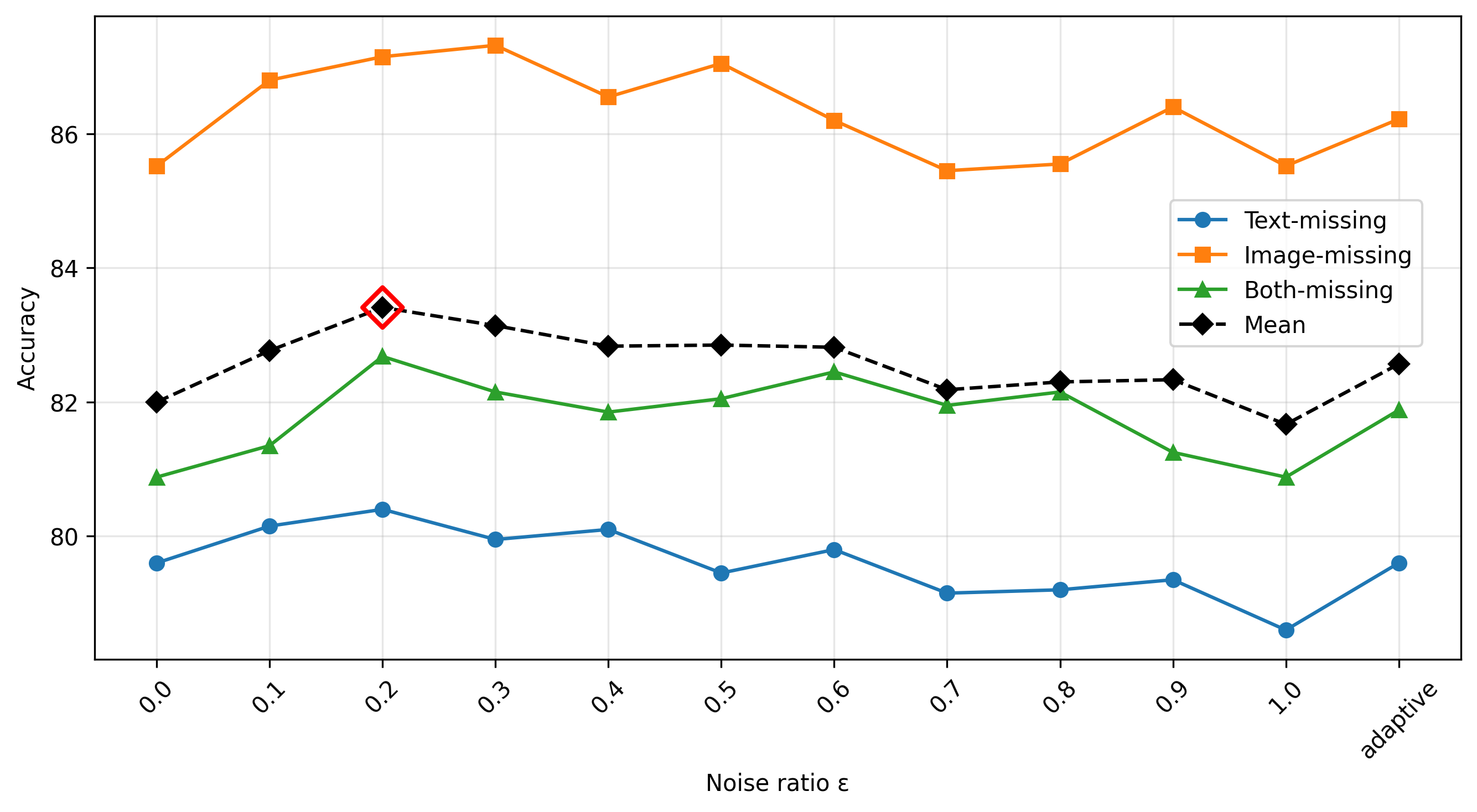}
	\caption{Ablation study on the value of noise factor $\epsilon_n$.}
	\label{fig_noise}
	\vspace{-0.1cm}
\end{figure}

The results in Table~\ref{tab:food101_ablation} show that each module in REP contributes steadily to performance improvements. Dynamic initialization provides a clear initial gain, while introducing dual private–shared buffers further enhances robustness under all missing-modality conditions. Adding the replay mechanism yields the largest improvement, highlighting its role in recovering early-layer information and stabilizing predictions when inputs are incomplete.

Table~\ref{tab:food101_ablation_part2} further examines the design choices within REP. Removing either the private buffer leads to consistent performance drops, indicating that decoupling is necessary. Finally, replacing Gaussian noise in initialization with uniform or Laplace noise results in slightly lower accuracy, confirming that the proposed initialization strategy is effective and well aligned with the learning dynamics of REP.

In addition, we investigate the influence of the noise ratio $\epsilon_n$. As shown in the Figure \ref{fig_noise}, although $\epsilon_n = 0.3$ yields the best performance in image-missing, the overall average performance is highest when $\epsilon_n = 0.2$. Therefore, we adopt $\epsilon_n = 0.2$ as the default setting for all experiments.

\textbf{Buffer settings and Efficiency analysis.} Large buffer sizes inevitably increase both computational overhead and parameter count, while offering limited additional performance benefits. To better understand this trade-off, we systematically vary the prompt length and buffer depth. 
As shown in Figure~\ref{fig5}, the model achieves the best performance when the depth is set to 6 and the prompt length to 36, beyond which the gains saturate or even slightly decline.

To further ensure fairness when comparing with existing methods, we additionally control the number of learnable parameters by adjusting the prompt length. The results are summarized in Table~\ref{tab:memory}.

\begin{table}[!htbp]
	\centering
	\caption{Efficiency analysis of REP on Food-101}
	\label{tab:memory}
	\resizebox{0.6\columnwidth}{!}{
		\begin{tabular}{lcccccc}
			\toprule
			\textbf{Method} & \multicolumn{6}{c}{\textbf{Params (M)}} \\
			\cmidrule{2-7}  
			& \footnotesize 0.08 & \footnotesize 0.64 & \footnotesize 1.12 & \footnotesize 1.6 & \footnotesize 2.08 & \footnotesize 2.56 \\
			\midrule
			MAP & 77.52 & 78.14 & 78.59 & 79.08 & 79.24 & 79.11 \\
			DCP & 80.04 & 80.89 & 80.82 & 81.87 & 81.80 & 81.76 \\
			Ours & 81.16 & 82.04 & 82.23 & \textbf{82.68} & 81.85 & 81.64 \\
			\bottomrule
	\end{tabular}}
	
\end{table}

Under the same parameters increasement, REP consistently outperforms MAP and DCP. Moreover, it achieves strong robustness in missing-modality scenarios without requiring a large increase in parameters. These findings indicate that REP provides a favorable balance between efficiency and performance.

\section{Conclusion}

REP is designed to address the core challenge of information decay in missing-modality scenarios. By introducing dynamic initialization, private–shared feature decomposition, and cross-layer replay, REP effectively preserves early-layer representations and compensates for incomplete inputs. The experimental results across VL and VLA benchmarks confirm that REP successfully mitigates the limitations outlined in our motivation and delivers consistent robustness improvements.

\bibliographystyle{unsrt}  
\bibliography{references}  

\clearpage
\setcounter{page}{1}

\section{Appendix}
\subsection{More Results}
\label{A.A}

This section presents additional results of REP on vision-language (VL) tasks. We conduct evaluations on three multimodal benchmark datasets—Food-101, MMIMDB, and Hateful Memes—to verify REP’s robustness and generalization under various modality-missing scenarios.

\textbf{Complete Training Setup.} All models are trained on fully available image-text pairs. At test time, we gradually increase the missing rate $\eta$ from 0\%, 30\%, 50\%, 70\% to 90\% to simulate different levels of information loss. We use the pre-trained CLIP as the baseline, and compare REP against state-of-the-art methods including MAP, MMP, and DCP.

\begin{figure*}[ht]
	\centering
	\includegraphics[width=0.9\textwidth]{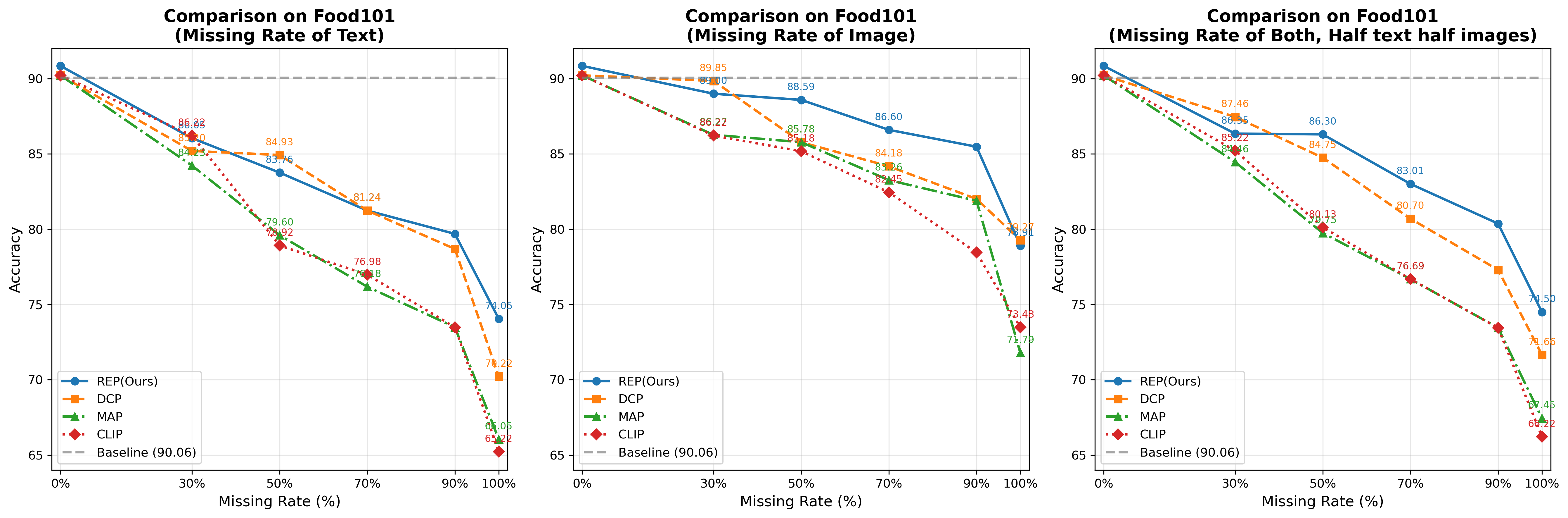}
	\caption{Evaluation on the Food-101 dataset under full training setup. All models are trained with 100\% image and text data. The three subplots show results under text-missing, image-missing, and both-modality-missing settings, respectively.}
	\label{fig:4}
\end{figure*}

As shown in Figure~\ref{fig:4}, under no missing conditions, all methods perform comparably (CLIP baseline: 90.06). However, as the missing rate increases, REP exhibits significantly slower performance degradation. For example, under 30\% text missing, REP achieves an accuracy of 86.05, outperforming DCP (85.20), MAP (83.94), and CLIP (82.93) by 1.8\%–3.12\%. Even under 50\% image missing, REP still maintains 88.56 accuracy, almost equal to the no-missing baseline.

In extreme missing scenarios ($\eta \geq 70\%$), REP shows even more pronounced advantages. For example, with 70\% missing, REP achieves 83.01 accuracy, surpassing MAP (75.66) and CLIP (76.69) by 7.35\% and 6.32\%, respectively. Even with 100\% modality missing, REP still achieves 72.05 (image-only) and 78.91 (text-only), significantly outperforming all baselines.

These results confirm that REP, trained on complete data, effectively captures both single-modality and cross-modality signals, enabling strong robustness under modality-missing conditions. It prevents the performance collapse commonly observed in traditional models relying solely on a single modality.

\vspace{0.5em}
\textbf{Cross-Dataset Generalization.} To further verify REP’s robustness in broader settings, we evaluate it on MMIMDB and Hateful Memes, two widely used multimodal benchmarks.

\begin{figure*}[htbp]
	\centering
	\includegraphics[width=0.9\textwidth]{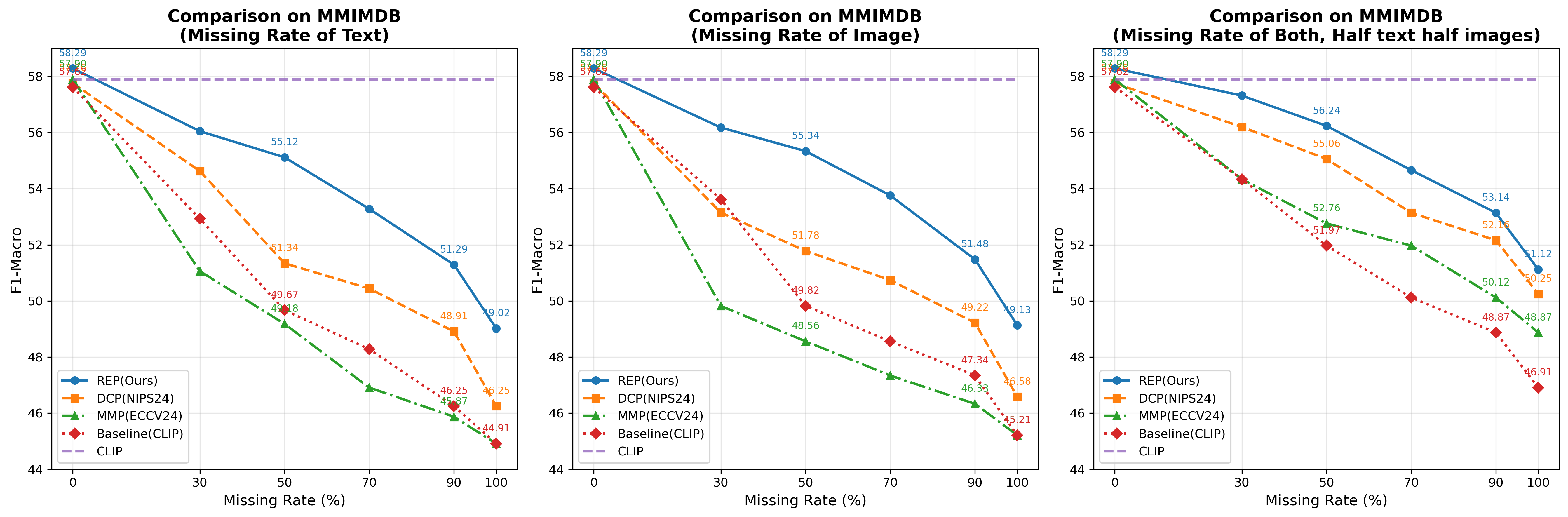}
	\caption{Evaluation on MMIMDB under full training. All models are trained with complete data and evaluated under varying missing rates. }
	\label{fig:7}
\end{figure*}

\begin{figure*}[htbp]
	\centering
	\includegraphics[width=0.9\textwidth]{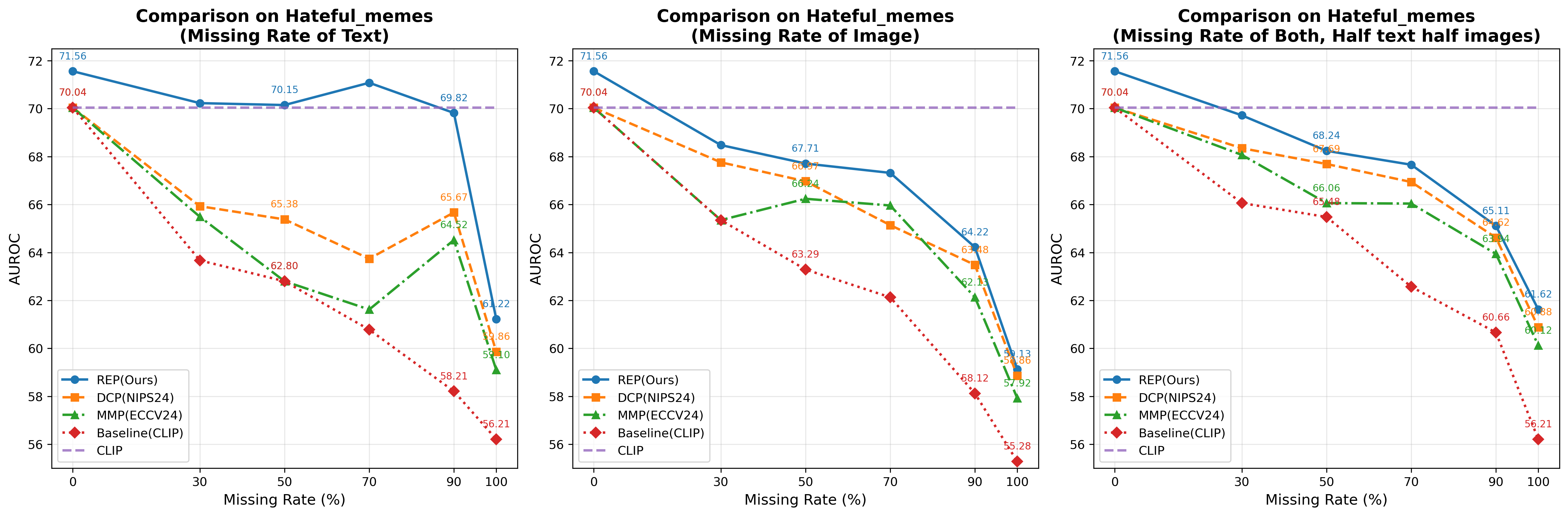}
	\caption{Evaluation on Hateful Memes under full training. All models are trained with complete data and evaluated under varying missing rates.}
	\label{fig:8}
\end{figure*}
As shown in Figure~\ref{fig:7} and ~\ref{fig:8}, REP consistently outperforms prior methods across both datasets, especially under severe or single-modality-missing conditions.

On MMIMDB, REP achieves 58.29 F1-Macro with no missing, slightly outperforming the baseline (57.76). With 30\% text missing, REP achieves 56.05, outperforming DCP (55.20), MMP (54.94), and the baseline (52.93). Even under 100\% text missing, REP maintains 49.02 F1-Macro, outperforming others by 2.77\%–4.11\%.

On Hateful Memes, REP achieves 68.48 AUROC with 30\% image missing, outperforming DCP (67.56) and baseline (65.35). Under 100\% image missing, REP still reaches 59.13 AUROC, surpassing the baseline (55.28) by 3.85\% and outperforming MMP (57.92) and DCP (58.86).

When both modalities are heavily missing (e.g., 90\% each), REP continues to lead with 65.11 AUROC, outperforming DCP (64.62) and MMP (63.94). In text-only and image-only testing scenarios, REP still shows clear advantages. For instance, on Hateful Memes, REP achieves 61.22 (text-only) and 59.13 (image-only) AUROC, outperforming the baseline (56.21 and 55.28).

Furthermore, REP demonstrates significantly slower performance degradation compared to baselines. On MMIMDB, increasing the text missing rate from 0\% to 100\% results in only a 9.27-point drop in REP's F1-Macro (from 58.29 to 49.02), while the baseline drops by 12.85 points. Similarly, on Hateful Memes, REP’s AUROC decreases by 12.43 (from 71.56 to 59.13), while the baseline drops by 14.76.

\subsection{More Ablation on Buffer}
\label{A.B}

We conduct ablation studies to analyze how the buffer length \( L \) and depth \( D \) affect model performance. Experiments are performed on MMIMDB and Hateful Memes under 70\% missing rate, covering three scenarios: text missing, image missing, and both-missing. Figure~\ref{fig:8} reports the results.

\begin{figure*}[t]
	\centering
	\includegraphics[width=\textwidth]{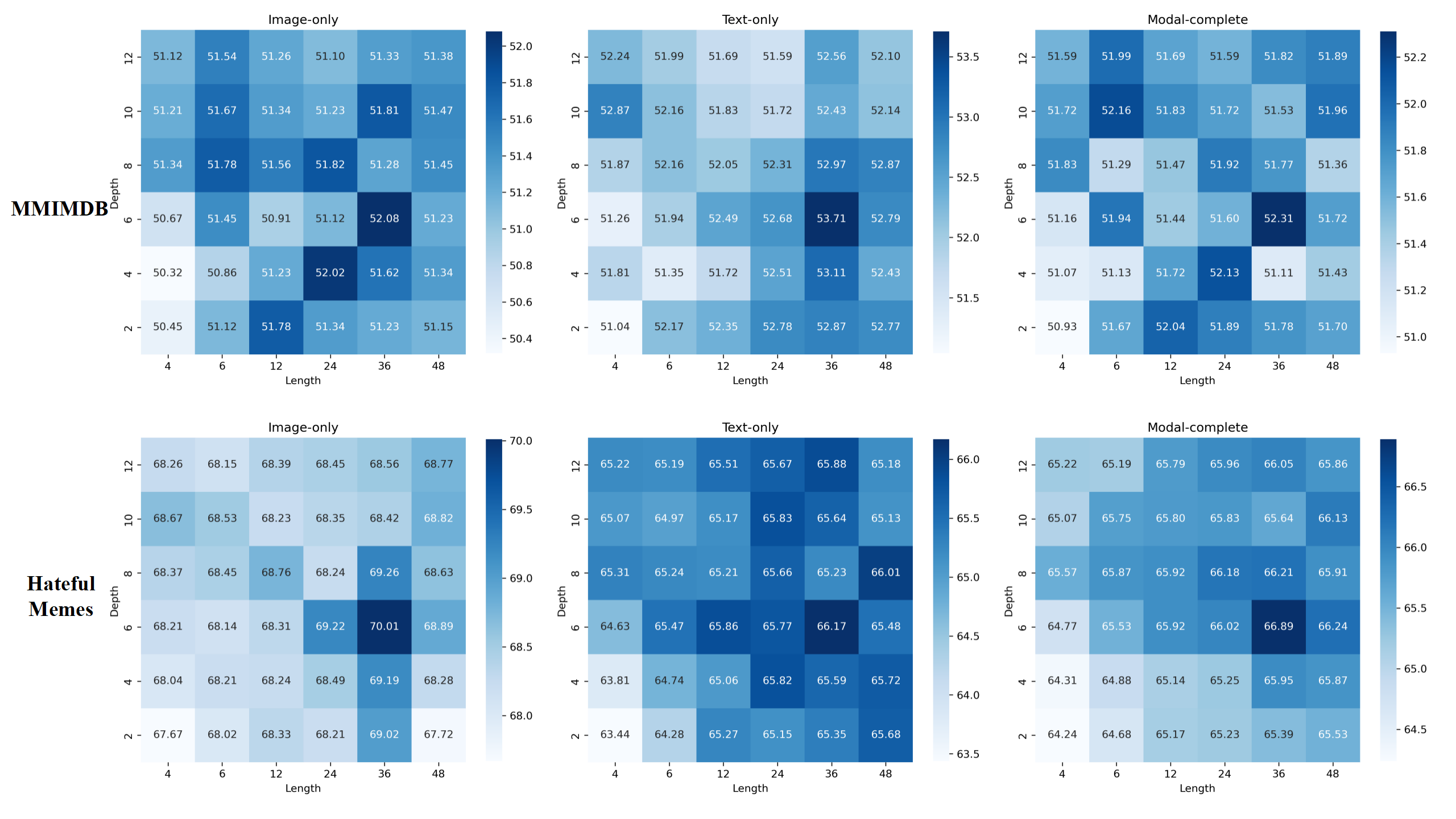}
	\caption{Ablation on buffer length and depth. Top: MMIMDB. Bottom: Hateful Memes. Each column corresponds to a different missing setting.}
	\label{fig8}
\end{figure*}

The optimal performance is generally achieved when \( L = 36 \) and \( D = 6 \). In MMIMDB’s text-missing setting, REP achieves 53.71 at \( L=36, D=6 \), clearly higher than nearby configurations (e.g., 52.68 at \( L=24 \)). Similarly, for Hateful Memes in the full-modality setting, performance peaks at 66.89 with \( L=36 \), but drops to 66.24 when \( L=48 \).

As \( L \) or \( D \) increases further, the performance shows diminishing returns or even degradation, likely due to feature redundancy or computation overhead. For instance, increasing \( D \) from 6 to 12 at \( L=36 \) on Hateful Memes degrades performance from 66.89 to 66.05. This suggests that overly deep or long buffers may harm aggregation.

\subsection{Comparison with Multimodal Large Models}
\label{A.C}

We compare REP with open-source multimodal LLMs: LLaVA-v1.5-7B and Qwen2-VL-7B-Instruct, under missing-modality settings on Hateful Memes. Results are shown in Table~\ref{tab:llm}.

\begin{table}[ht]
	\centering
	\caption{Performance comparison on Hateful Memes with different modality-missing conditions.}
	\label{tab:llm}
	\resizebox{0.6\columnwidth}{!}{
		\begin{tabular}{lccccc}
			\toprule
			\textbf{Method} & \textbf{Image} & \textbf{Text} & \textbf{Accuracy} & \textbf{AUROC} \\
			\midrule
			DCP & 100\% & 30\% & 69.89 & 64.12 \\
			& 30\% & 100\% & 67.85 & 65.48 \\
			& 65\% & 65\% & 68.92 & 66.08 \\
			\midrule
			REP & 100\% & 30\% & 71.25 & \textbf{70.01} \\
			& 30\% & 100\% & 68.22 & 66.17 \\
			& 65\% & 65\% & \textbf{70.41} & \textbf{66.89} \\
			\midrule
			LLaVA-7B & 100\% & 30\% & 55.44 & 50.32 \\
			& 30\% & 100\% & 63.57 & 57.25 \\
			& 65\% & 65\% & 60.48 & 58.12 \\
			\midrule
			Qwen2-7B & 100\% & 30\% & 60.57 & 55.92 \\
			& 30\% & 100\% & 65.98 & 62.44 \\
			& 65\% & 65\% & 62.72 & 60.25 \\
			\bottomrule
		\end{tabular}
	}
\end{table}

REP significantly outperforms both LLMs in all cases. Notably, under text-missing conditions, LLaVA and Qwen2 drop to near-random performance (AUROC 50.32 and 55.92), while REP maintains 70.01 AUROC. Considering the huge tuning cost of MLLMs, REP offers a lightweight and robust alternative.
\end{document}